\documentclass[11pt,a4paper]{article}
\usepackage{hyperref}
\usepackage[hyperref]{acl2019}
\usepackage{times}
\usepackage{latexsym}
\usepackage{graphicx}
\usepackage{amsmath}
\usepackage{multirow}
\usepackage{url}

\aclfinalcopy 
\def\aclpaperid{5} 

\newcommand{\afil}[1]{\textsuperscript{#1}}

\title{Conversational Response Re-ranking Based on Event Causality\\and Role Factored Tensor Event Embedding}

\author{Shohei Tanaka\afil{1},
Koichiro Yoshino\afil{1}\afil{,}\afil{2}, 
Katsuhito Sudoh\afil{1},
Satoshi Nakamura\afil{1}
\\[3pt]
\afil{1} Nara Institute of Science and Technology\\
\quad \quad \afil{2} PRESTO, Japan Science and Technology Agency\\
\texttt{\{takana.shohei.tj7, koichiro, sudoh, s-nakamura\}@is.naist.jp}
}

\date{}

\begin{document}
\maketitle
\begin{abstract}
We propose a novel method for selecting coherent and diverse responses for a given dialogue context.
The proposed method re-ranks response candidates generated from conversational models by using event causality relations between events in a dialogue history and response candidates (e.g., ``be stressed out'' precedes ``relieve stress'').
We use distributed event representation based on the Role Factored Tensor Model for a robust matching of event causality relations due to limited event causality knowledge of the system.
Experimental results showed that the proposed method improved coherency and dialogue continuity of system responses.
\end{abstract}


\section{Introduction}
While a variety of dialogue models such as the neural conversational model (NCM) \cite{ncm} have been researched widely,
such dialogue models often generate simple and dull responses due to the limitation of their ability to take dialogue context into account.
It is very difficult for these models to generate coherent responses to a dialogue history.
We tackle this problem with a new architecture by incorporating event causality relations between response candidates and a dialogue history.
Typical event causality relations are cause-effect relations between two events, such as ``be stressed out" precedes ``relieve stress."
In this paper, event causality relations are defined that an effect event is likely to happen after a corresponding cause event happens \cite{ecdic, ecdic2}.
Event causality relations have been used in why-question answering systems to focus on causalities between questions and answers \cite{intra, semisuper, mcnn-ca}.
It is also reported that a conversational model using event causality relations can generate diverse and coherent responses \cite{fujita}.
However, the relation between dialogue continuity and the coherency of system responses is still an underlying problem.\par
In this paper, we propose a novel method to select an appropriate response from response candidates generated by NCMs.
We define a score for re-ranking to select a response that has an event causality relation to a dialogue history.
Re-ranking effectively improves response reliability in language generation tasks such as why-question answering and dialogue systems \cite{intra, jansen, bogdanova, ohmura}.
We used event causality pairs extracted from a large-scale corpus \cite{ecdic, ecdic2}.
We also use distributed event representation based on the Role Factored Tensor Model (RFTM) \cite{evnttnsr} to realize a robust matching of event causality relations, even if these causalities are not included in the extracted event causality pairs.
In human and automatic evaluations, the proposed method outperformed conventional methods in selecting coherent and diverse responses.

\begin{figure*}[t]
  \begin{center}
  \includegraphics[width=15cm]{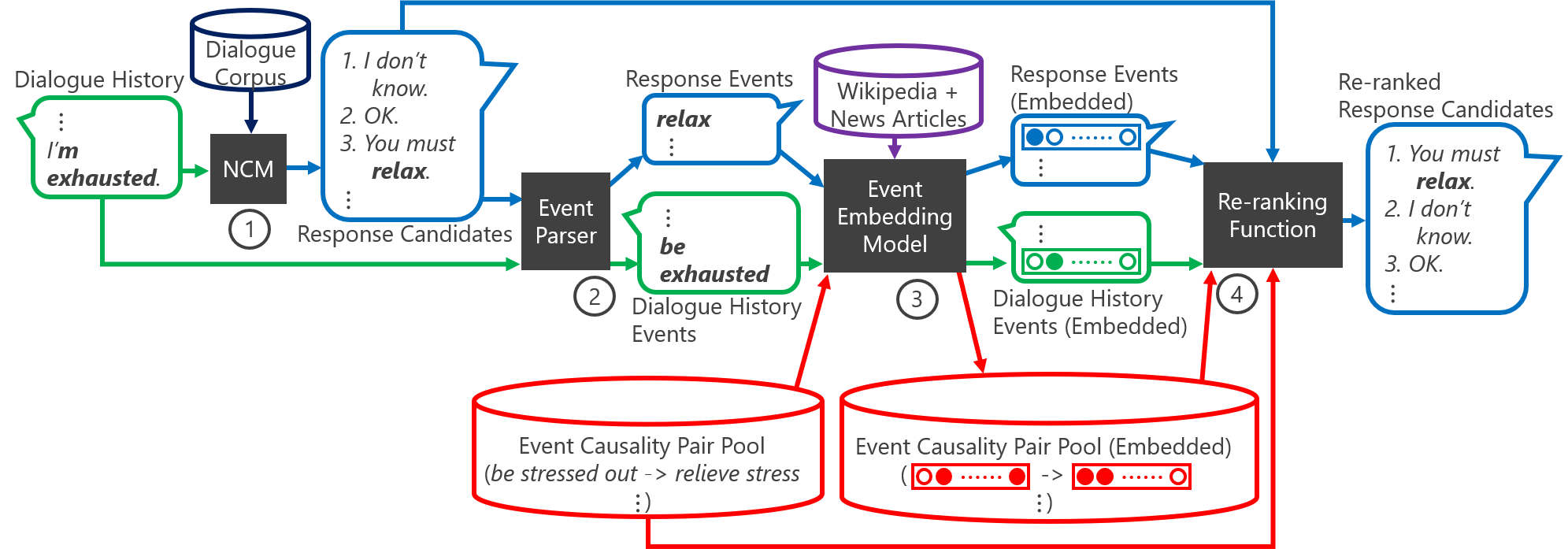}
  \end{center}
  \caption{Neural conversational model$+$re-ranking using event causality; 
  a response that has an event causality relation (``{\em be exhausted}'' $\rightarrow$ ``{\em relax}'') to the dialogue history is selected by the re-ranking.}
  \label{fig:model}
\end{figure*}

\begin{table*}[t]
  \begin{center}
  \small
  \begin{tabular}{|c|c|c|c|c|}
  \hline
  predicate 1 & argument 1 & predicate 2 & argument 2 & $lift$ \\
  \hline
  \hline
  be stressed out & - & relieve & stress & \multicolumn{1}{r|}{10.02} \\
  \hline
  \end{tabular}
  \normalsize
  \end{center}
  \caption{Example of event causality relations included in event causality pairs}
  \label{tab:ecdic}
\end{table*}


\section{Response Re-ranking Using Event Causality Relations}
Figure \ref{fig:model} shows an overview of the proposed method.
The process consists of four parts.
First, $N$-best response candidates are generated from an NCM given a dialogue history (Figure \ref{fig:model} \textcircled{\scriptsize 1}; Section \ref{sec:ncm}).
Then, events (predicate-argument structures) are extracted by an event parser from both the dialogue history and the response candidates (Figure \ref{fig:model} \textcircled{\scriptsize 2}).
We used Kurohashi Nagao Parser (KNP)\footnote{http://nlp.ist.i.kyoto-u.ac.jp/?KNP} \cite{knp, knp2} as the event parser.
Next, the extracted events are converted to distributed event representations by an event embedding model (Figure \ref{fig:model} \textcircled{\scriptsize 3}; Section \ref{sec:rftm}).
Events in event causality pairs are also converted to distributed representations to calculate similarities.
The RFTM is used for the embedding.
Finally, response candidates are re-ranked (Figure \ref{fig:model} \textcircled{\scriptsize 4}; Section \ref{sec:ecdic}, \ref{sec:matching}).
We describe these components in more detail below.

\subsection{Neural Conversational Model (NCM)\label{sec:ncm}}
NCM learns a mapping between input and output word sequences by using recurrent neural networks (RNNs).
NCMs can generate $N$-best response candidates by using beam search or sampling \cite{lennorm}.

\subsection{Event Causality Pairs\label{sec:ecdic}}
The proposed method uses event causality pairs.
Events in a pair, which have cause-effect relations, are extracted from a large-scale corpus on the basis of co-occurring statistics and case frames \cite{ecdic, ecdic2}.
420,000 entries are extracted from 1.6 billion texts: each entry consists of information denoted in Table \ref{tab:ecdic}. 
``predicate 1'' and ``argument 1'' are components of a cause event, and ``predicate 2'' and ``argument 2'' are components of an effect event.
Each event consists of a predicate and arguments.
The predicate is required, and the argument is optional.
We used arguments that have the following roles: nominative, accusative, dative, instrumental, and locative cases.
$lift$ is the mutual information score between two events, which indicates the strength of the causality relation.
Using $lift$, we propose a score for re-ranking as,
\begin{align}
  score &= \max_{<{\rm e}_h, {\rm e}_r>} \frac{\log_2 p}{\bigl(\log_2 lift({\rm e}_h, {\rm e}_r)\bigr)^{\lambda}}. \label{eq:noemb}
\end{align}
\noindent
$p$ is the posterior probability of the response candidate provided by NCM.
$\lambda$ is a hyper parameter to decide the weight of event causality relations.
$lift({\rm e}_h, {\rm e}_r)$ is the $lift$ score between an event ${\rm e}_h$ in the dialogue history, and an event ${\rm e}_r$ in the response candidate, which is equal to 2 if the pair does not appear in the extracted event causality pair pool.
Note that $lift({\rm e}_h, {\rm e}_r)$ is log-scaled because it has a wide range of values $(10 < lift({\rm e}_h, {\rm e}_r) < 10,000)$.
In the case where more than one event causality relations are recognized between the dialogue history and the response candidate, the score of the candidate is determined by the relation with the highest $lift({\rm e}_h, {\rm e}_r)$.
We call this model ``Re-ranking.''

\subsection{Distributed Event Representation Based on Role Factored Tensor Model (RFTM)\label{sec:rftm}}
It is difficult to determine event causality relations by using only the pairs observed in an actual corpus.
Therefore, we introduce a distributed event representation to improve the robustness of matching events in a dialogue with those in the event causality pair pool.
Any events are embedded into fixed length vectors to calculate their similarities.\par
\begin{figure}[t]
  \begin{center}
  \includegraphics[width=7cm]{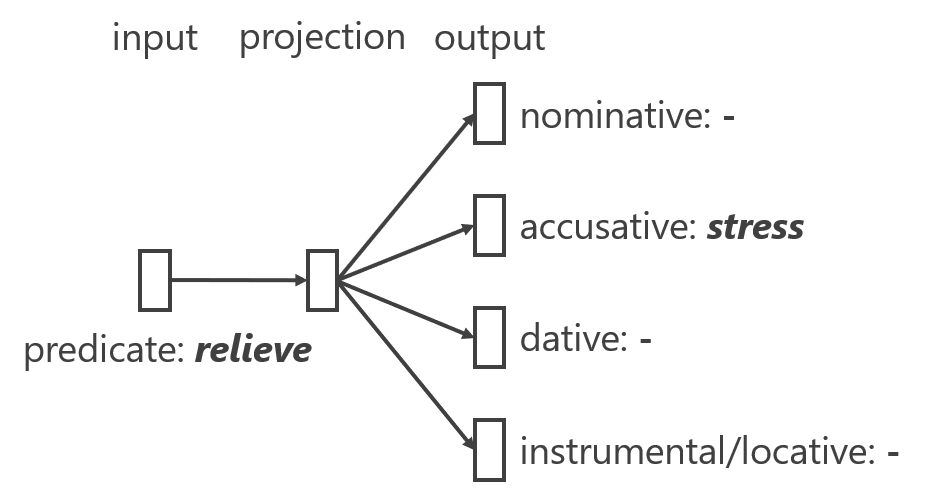}
  \end{center}
  \caption{Model architecture of predicate embedding}
  \label{fig:preemb}
\end{figure}

We define an event with a single predicate or a pair of a predicate and arguments.
Argument $a$ of an event is embedded into vector as ${\rm v}_{a}$ by using Skip-gram \cite{word2vec1, word2vec2, word2vec3}.
Predicate $p$ of an event is embedded into vector as ${\rm v}_{p}$ by using predicate embedding which is based on case-unit Skip-gram.
Figure \ref{fig:preemb} shows the model architecture of predicate embedding.
The model learns predicate vector representations which are good at predicting its arguments.
To get an event embedding for the pair of ${\rm v}_{p}$ and ${\rm v}_{a}$, we propose to use RFTM, which was proposed by \citet{evnttnsr}.
The RFTM embeds a predicate and its arguments into vector ${\rm e}$ as,
\begin{align}
  {\rm e} &= \sum_{a} W_{a} T({\rm v}_{p}, {\rm v}_{a}). \label{eq:rft}
\end{align}
\noindent
The relation of a predicate and its arguments is computed using a 3D tensor $T$ and matrices $W_{a}$.
If the event has no arguments, ${\rm e}$ is substituted by ${\rm v}_{p}$.
The RFTM is trained to predict an event sequence; thus it can represent the meaning of the event in a particular context.
\subsection{Event Causality Relation Matching Based on Distributed Event Representation\label{sec:matching}}
\begin{figure}[t]
  \begin{center}
  \includegraphics[width=7cm]{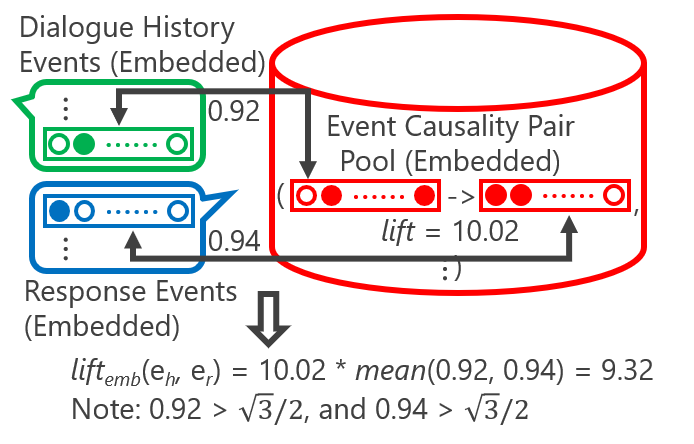}
  \end{center}
  \vspace{-3mm}
  \caption{Event causality relation matching; 
  the $lift$ of the event causality relation in which ``{\em be exhausted}'' precedes ``{\em relax},'' is calculated from the $lift$ of the most similar event causality relation where ``{\em be stressed out}'' precedes ``{\em relieve stress}.''}
  \label{fig:matching}
\end{figure}

Figure \ref{fig:matching} illustrates the process of matching events on the basis of distributed event representation.
Given an event pair from a response candidate and a dialogue history, the proposed method finds an event causality pair that has the highest cosine similarity from the pool.
$lift$ score, strength of the event causality relation, is extended as,
\begin{align}
  &lift_{emb}({\rm e}_h, {\rm e}_r) =\nonumber \\
  &lift({\rm e}_c, {\rm e}_e) * mean \bigl(sim({\rm e}_h, {\rm e}_c), sim({\rm e}_r, {\rm e}_e)\bigr). \label{eq:lift}
\end{align}
\noindent
${\rm e}_h$ is an event in the dialogue history, ${\rm e}_r$ is an event in the response candidate.
${\rm e}_c$ and ${\rm e}_e$ are respectively a cause and an effect event of an event causality pair.
We also calculate the score for the case in which the cause and effect events are exchanged to deal with the inverse case.
Note that both $sim$ values have a threshold to prevent over-generalization.
The threshold was empirically decided as $\sqrt{3}/2$.
Replacing $lift({\rm e}_h, {\rm e}_r)$ in Eq. \eqref{eq:noemb} with $lift_{emb}({\rm e}_h, {\rm e}_r)$, the score using distributed event representation is defined as,
\begin{align}
  score &= \max_{<{\rm e}_h, {\rm e}_r>} \frac{\log_2 p}{\bigl(\log_2 lift_{emb}({\rm e}_h, {\rm e}_r)\bigr)^{\lambda}}. \label{eq:emb}
\end{align}
\noindent
We call this model ``Re-ranking (emb).''

\section{Experiments\label{sec:exp}}
\begin{table}[t]
  \begin{center}
  \small
  \begin{tabular}{|l||r|r|}
  \hline
  & \multicolumn{1}{l|}{${\rm Ave. dist\mathchar`-1}$} & \multicolumn{1}{l|}{${\rm Ave. dist\mathchar`-2}$} \\
  \hline
  \hline
  EncDec & \mbox{\boldmath $0.44$} & \mbox{\boldmath $0.56$} \\
  \hline
  HRED & 0.33 & 0.42 \\
  \hline
  \end{tabular}
  \normalsize
  \end{center}
  \vspace{-2mm}
  \caption{Diversity of $N$-best Response Candidates}
  \label{tab:n-dist}
  \vspace{-3mm}
\end{table}
\begin{table*}[t]
  \begin{center}
  \scriptsize
  \begin{tabular}{|l|l|l||r|r|r|r|r|r|r|r|}
  \hline
  \multicolumn{3}{|l||}{Method}
  & \multicolumn{8}{l|}{Evaluation} \\
  \hline
  NCM
  & history
  & re-ranking
  & \multicolumn{1}{l|}{re-ranked (\%)}
  & \multicolumn{1}{l|}{BLEU}
  & \multicolumn{1}{l|}{NIST}
  & \multicolumn{1}{l|}{extrema}
  & \multicolumn{1}{l|}{dist-1}
  & \multicolumn{1}{l|}{dist-2}
  & \multicolumn{1}{l|}{PMI}
  & \multicolumn{1}{l|}{length} \\
  \hline
  \hline
  reference & - & - & - & - & - & - & 0.06 & 0.40 & 1.86 & 21.43 \\
  \hline
  \hline
  EncDec & - & 1-best & - & 1.12 & 1.19 & \mbox{\boldmath $0.42$} & 0.06 & 0.18 & 1.77 & 15.55 \\
  \hline
  EncDec & 1 & Re-ranking & 4,016 (7.90) & 1.10 & 1.18 & \mbox{\boldmath $0.42$} & 0.06 & 0.19 & 1.78 & 15.52 \\
  \hline
  EncDec & 1 & Re-ranking (emb) & 29,343 (57.71) & 1.02 & 1.07 & 0.40 & 0.06 & 0.20 & 1.77 & 15.64 \\
  \hline
  EncDec & 5 & Re-ranking & 6,469 (12.72) & 1.09 & 1.17 & \mbox{\boldmath $0.42$} & 0.06 & 0.19 & 1.78 & 15.50 \\
  \hline
  EncDec & 5 & Re-ranking (emb) & 35,284 (69.39) & 1.00 & 1.04 & 0.39 & \mbox{\boldmath $0.07$} & \mbox{\boldmath $0.21$} & 1.77 & 15.66 \\
  \hline
  HRED & - & 1-best & - & \mbox{\boldmath $1.34$} & \mbox{\boldmath $2.74$} & \mbox{\boldmath $0.42$} & \mbox{\boldmath $0.07$} & 0.20 & 1.84 & 35.05 \\
  \hline
  HRED & 1 & Re-ranking & 3,671 (7.22) & 1.33 & \mbox{\boldmath $2.74$} & \mbox{\boldmath $0.42$} & 0.06 & 0.20 & 1.84 & 35.20 \\
  \hline
  HRED & 1 & Re-ranking (emb) & 30,992 (60.95) & 1.28 & \mbox{\boldmath $2.74$} & 0.41 & 0.06 & 0.20 & \mbox{\boldmath $1.86$} & 34.80 \\
  \hline
  HRED & 5 & Re-ranking & 6,231 (12.25) & 1.33 & 2.73 & \mbox{\boldmath $0.42$} & 0.06 & 0.20 & 1.84 & \mbox{\boldmath $35.30$} \\
  \hline
  HRED & 5 & Re-ranking (emb) & \mbox{\boldmath $36,373 (71.53)$} & 1.28 & \mbox{\boldmath $2.74$} & 0.41 & 0.06 & 0.20 & \mbox{\boldmath $1.86$} & 34.60 \\
  \hline
  \end{tabular}
  \normalsize
  \end{center}
  \vspace{-2mm}
  \caption{Comparison results before and after re-ranking}
  \label{tab:automatic}
  \vspace{-3mm}
\end{table*}
%
We conducted automatic and human evaluations to compare responses with and without the re-ranking.
We evaluated our proposed re-ranking method on a conventional Encoder-Decoder with Attention (EncDec) model \cite{attn, attn2} and a Hierarchical Recurrent Encoder-Decoder (HRED) model \cite{hred, hred2}.
While HRED tries to generate more coherent responses to dialogue context than a simple Encoder-Decoder, the diversity of responses is small due to context constraints.
\par
We used the Japanese data from a Wikipedia dump for training Skip-gram and predicate word embeddings of RFTM, and the {\em Maichichi} newspaper dataset 2017\footnote{http://www.nichigai.co.jp/sales/mainichi/mainichi-data.html} for training RFTM.
We collected 2,632,114 dialogues from Japanese micro blogs (Twitter) to train and test the dialogue models.
The average dialogue turn was 21.99, and the average utterance length was 22.08 words.
We removed emoticons from utterances to reduce vocabulary size and accelerate the training.
The dialogue corpus was split into 2,509,836, 63,308, and 58,970 dialogues as training, validation, and testing data, respectively.
%
\subsection{Model Settings}
The hidden unit size of Skip-gram \cite{word2vec1, word2vec2, word2vec3}, predicate embedding, and RFTM \cite{evnttnsr} was 100.
We used gated recurrent units (GRUs) \cite{gru, gru2} whose number of layers was 2 and hidden unit size was 256, for the encoder and decoder of the NCMs.
The batch size was 100, the dropout probability was 0.1, and the teacher forcing rate was 1.0.
We used Adam \cite{adam} as the optimizer.
The gradient clipping was 50, the learning rate for the encoder and the context RNN of HRED was $1\mathrm{e}^{-4}$, and the learning rate for the decoder was $5\mathrm{e}^{-4}$.
The loss function was inverse token frequency (ITF) loss \cite{itf}.
We used sentencepiece \cite{sentpc} as the tokenizer, and the vocabulary size was 32,000.
These settings were the same in all models.\par
Repetitive suppression \cite{itf} and length normalization \cite{lennorm} were used at the decoding step.
Finally, $\lambda$ of Eq. \eqref{eq:noemb} and Eq. \eqref{eq:emb} was set to 1.0. 
\subsection{Diversity of Beam Search}
%
We investigated internal diversity of $N$-best response candidates generated from each dialogue model.
It is expected that the higher diversity is, the more effective re-ranking is.
Hence, we evaluated diversity on the test data by dist-1, 2 \cite{dist}.
Beam width was set to 20; it is same in the following experiments.\par
The result is shown in Table \ref{tab:n-dist}: ${\rm Ave. dists}$ are averages of dist computed internal $N$-best response candidates.
The diversity of EncDec is higher than that of HRED.
%
\subsection{Comparison in Automatic Metrics}
%
Table \ref{tab:automatic} shows the results of our evaluation using automatic metrics.
We compared the results by referring to the ratio of responses different from the without re-ranking method (``re-ranked''), bilingual evaluation understudy (BLEU) \cite{bleu}, NIST \cite{nist}, and vector extrema \cite{extrema} (``extrema'') score.
NIST is based on BLEU, but heavily weights less frequent N-grams to focus on content words.
Vector extrema computes cosine similarity between sentence vectors of a reference and a generated response from a model.
Each sentence vector ${\rm e}_{s}$ is computed by taking extrema of Skip-gram word vectors ${\rm e}_{w}$ in each dimension $d$ as,
\begin{align}
  e_{sd} &=& \begin{cases}
  \max_{w \in s} e_{wd} & {\rm if} \hspace{2mm} e_{wd} > |\min_{w' \in s} e_{w'd}| \\
  \min_{w \in s} e_{wd} & {\rm otherwise}
  \end{cases}. \label{eq:extrema}
\end{align}
$e_{sd}$ and $e_{wd}$ are the $d$th dimensions of ${\rm e}_{s}$ and ${\rm e}_{w}$ respectively.
Additionally, we evaluated dist \cite{dist}, Pointwise Mutual Information (PMI) \cite{pmi}, and average response length (``length'').
Dist and PMI are used to evaluate diversity and coherency respectively.
PMI between a response and a dialogue history is defined as,
\begin{align}
  {\rm PMI} &=& \frac{1}{|response|} \sum^{|response|}_{wr} \max_{wh} {\rm PMI}(wr, wh). \label{eq:pmi}
\end{align}
\noindent
$wr$ and $wh$ are words in the response and the dialogue history respectively.
Each method used a specific NCM, a range of dialogue history used for re-ranking, and re-ranking method.
Methods with ``1-best'' used neither re-ranking and event embedding.
Those with ``Re-ranking'' used re-ranking but did not use event embedding.
Those with ``Re-ranking (emb)'' used both the re-ranking and the proposed event embedding method.
\par
Re-ranking lowered scores of the similarity to reference: BLEU, NIST, and extrema, because normal NCM models were trained to generate similar responses to the references, generated top 1 response before re-ranking should have the highest scores in those similarity metrics.
Dist-2 and PMI were improved by re-ranking.
This indicates that words in re-ranked responses are diverse and coherent to dialogue histories.
However, ratios of re-ranked responses were around 10\%; hence, the effect of re-ranking was limited.
By introducing the proposed event embedding method, the ratios of re-ranked responses improved drastically (Re-ranking vs. Re-ranking (emb)).
Moreover, the re-ranking models with event embedding have highest dist-1, dist-2, and PMI.
As the HRED models had higher BLEU, NIST, and PMI values than those of EncDec models in all re-ranking methods, we conducted a human evaluation by comparing HRED model-based systems.

\subsection{Human Evaluation}
\begin{table}[t]
  \begin{center}
    \small
    \begin{tabular}{|l|p{1.3cm}|p{1.3cm}|}
      \hline
      & word \hspace{0.3cm} coherency & dialogue continuity \\
      \hline
      \hline
      1-best & 28.62 & \mbox{\boldmath $40.84$} \\
      \hline
      Re-ranking & \mbox{\boldmath $33.91$} & 38.53 \\
      \hline
      neither & 37.47 & 20.62 \\
      \hline
    \end{tabular}
    \normalsize
  \end{center}
  \caption{1-best v.s. Re-ranking; \# dialogues is 100.}
  \label{tab:winrate1}
\end{table}
\begin{table}[t]
  \begin{center}
    \small
    \begin{tabular}{|l|p{1.3cm}|p{1.3cm}|}
      \hline
      & word \hspace{0.3cm} coherency & dialogue continuity \\
      \hline
      \hline
      1-best & \mbox{\boldmath $30.10$} & 35.50 \\
      \hline
      Re-ranking (emb) & 25.40 & \mbox{\boldmath $38.20$} \\
      \hline
      neither & 44.50 & 26.30 \\
      \hline
    \end{tabular}
    \normalsize
  \end{center}
  \caption{1-best v.s. Re-ranking (emb); \# dialogues is 100.}
  \label{tab:winrate2}
  \vspace{-3mm}
\end{table}
\begin{table}[t]
  \begin{center}
    \small
    \begin{tabular}{|l|p{1.3cm}|p{1.3cm}|}
      \hline
      & word \hspace{0.3cm} coherency & dialogue continuity \\
      \hline
      \hline
      Re-ranking & \mbox{\boldmath $23.70$} & 35.53 \\
      \hline
      Re-ranking (emb) & 22.91 & \mbox{\boldmath $35.65$} \\
      \hline
      neither & 55.39 & 28.83 \\
      \hline
    \end{tabular}
    \normalsize
  \end{center}
  \caption{Re-ranking v.s. Re-ranking (emb); \# dialogues is 100.}
  \label{tab:winrate3}
  \vspace{-3mm}
\end{table}
%
It is difficult to evaluate system performances only with automatic metrics \cite{hownotto}.
Hence, we compared a baseline model and our models in a human evaluation to confirm coherency and dialogue continuity of responses selected by our proposed methods.
We compared baseline HRED model with our proposed models, re-ranked without embedding and with embedding using the last five histories.
To reduce evaluators' workload, we used test data whose the number of user utterances is less than three, and removed dialogues which need external knowledge to evaluate.
We used crowdsourcing for the human evaluation.
Ten crowd-workers compared responses selected by two of three models in the following two subjective criteria.
The first one is ``which words in a response are more related to a dialogue history'' (word coherency), which indicates system response coherency to dialogue histories.
The second criterion is ``which response is easier to respond to'' (dialogue continuity), which indicates how much dialogue continuity system responses have.
We were inspired to make these criteria by those of the Alexa Prize \cite{alexa}.
\par
The results are shown in Table \ref{tab:winrate1}, \ref{tab:winrate2}, and \ref{tab:winrate3}.
Word coherency was improved by our model without embedding, but lowered by the model with embedding.
This is because workers acknowledged causality relations included in the event causality pair pool, but did not acknowledge generalized causalities with event embedding.
However, dialogue continuity was improved by the proposed re-ranking model with embedding, it is probably because the proposed model reduced the number of dull responses.
We need to investigate the better threshold in the event embedding to balance out the coherency and the continuity as the future work.
%
\section{Discussion}
We analyzed an adequacy of re-ranking using event causality relations.
Here are system response examples of our proposed method.
``()" indicates original Japanese sentences, ``[]" indicates event causality relations used for re-ranking, and ``$<>$" indicates responses before re-ranking.
All examples are translated from Japanese to English.
\\\\
\noindent
Conversation 1:
\begin{quote}
{\bf User 1:} Because of my fears, I have \textbf{been stressed out}. \\
({\em Mou fuan-na koto ga oosugite \textbf{sutoresu ga tama}tteku.}) \\
{\bf User 2 (System):} Are you OK? Don't \textbf{work too hard}. \\ß
({\em Daijobu desuka \textbf{muri shi}naide kudasaine}) \\
$[$work too hard $\rightarrow$ be stressed out \\
({\em muri wo suru} $\rightarrow$ {\em sutoresu ga tamaru})$]$ \\
$<$Are you OK? ({\em Daijobu desuka})$>$ \\
\end{quote}
Conversation 2:
\begin{quote}
{\bf User 1:} Someone, let's \textbf{go} to eat lunch around {\em Yayoi-ga-oka} tomorrow. Everyone is welcome. \\
({\em Dareka, ashita, yayoi-ga-oka shuhen de hiru-gohan tabe ni \textbf{iko} daredemo iiken onegai}) \\
{\bf User 2 :} Let's go. ({\em Ikimasuka}) \\
{\bf User 1:} Where will we go? \\
({\em Doko kui iku}) \\
{\bf User 2 (System):} Anywhere \textbf{near} {\em Shinjuku} station \\
({\em Shinjuku-eki no \textbf{chikaku}})\\
$[$be near $\rightarrow$ go ({\em chiaku} $\rightarrow$ {\em ikeru})$]$ \\
$<$Where are you? ({\em doko inno})$>$ \\
\end{quote}
\par
In above examples, appropriate event causality relations are used to select logical, coherent, and diverse responses.
However, we found that these cases are not majority.
Our method used inadequate event causality relations even if coherent responses were selected as a result (``Conversation 3'').
Responses selected by our method are sometimes unnatural and incoherent than those before re-ranking as ``Conversation 4'', ``Conversation 5'', or ``Conversation 6''.
\\\\
\noindent
Conversation 3:
\begin{quote}
{\bf User 1:} I can't \textbf{stop coughing}:\_( \\
({\em Hen na \textbf{seki ga tomara}nai...namida}) \\
{\bf User 2 (System):} Don't \textbf{catch a cold}. \\
({\em \textbf{kaze hika}nai dene}) \\
$[$catch cold $\rightarrow$ stop coughing \\
({\em kaze wo hiku} $\rightarrow$ {\em seki ga tomaru})$]$ \\
$<$Are you OK? ({\em daijobu-ka})$>$ \\
\end{quote}
Conversation 4:
\begin{quote}
{\bf User 1:} Our president who \textbf{can}not \textbf{drink alcohol} gives me beer... \\
({\em \textbf{Osake wo nome}nai shacho ga biru wo kureru...}) \\
{\bf User 2 (System):} I will \textbf{drink alcohol}.\\
({\em \textbf{Osake nomu}}.) \\
$[$drink alcohol $\rightarrow$ can drink alcohol\\
({\em osake wo nomu} $\rightarrow$ {\em osake wo nomeru})$]$ \\
$<$I see. ({\em wakaru})$>$ \\
\end{quote}
Conversation 5:
\begin{quote}
{\bf User 1:} Despite of \textbf{morning}, ambulances and police cars were noisy...  I was wondered what happened? \\
({\em \textbf{Asa} nanoni, kyu-kyu-sha to patoka no oto ga sugoi...doshitandaro}) \\
{\bf User 2 (System):} \textbf{Good morning}.
({\em \textbf{Ohayo}gozaimasu}) \\
$[$morning $\rightarrow$ good morning\\
({\em asa} $\rightarrow$ {\em ohayo})$]$ \\
$<$Are you OK? ({\em Daijobu desuka})$>$ \\
\end{quote}
Conversation 6:
\begin{quote}
{\bf User 1:} I, I can't get up... I'm gonna \textbf{take a day off}... I have a fever... \\
({\em O, okiagarenai... kaisha \textbf{yasumu}... netsu...}) \\
{\bf User 2 (System):} You must \textbf{be tired}.
({\em O\textbf{tsukare}sama desu}) \\
$[$be tired $\rightarrow$ take a day off\\
({\em tsukare} $\rightarrow$ {\em yasumu})$]$ \\
$<$Take care of yourself. ({\em Odaiji-ni})$>$ \\
\end{quote}
Considering the result of human evaluation and above examples, we hypothesized that our method have two problems in slecting appropriate event causality relations.
The first problem is that the event embedding over-generalized events (``Conversation 4'').
The causality in Conversation 4 (``drink alcohol'' precedes ``can drink alcohol'') is obtained by generalizing a causality that ``enter restaurant'' precedes ``order beer'', which is included in the event causality pair pool.
It is necessary to prevent over-generalization by improving the embedding architecture.
The second problem is that our method focuses on only word coherency, not response naturalness (``Conversation 5'' and ``Conversation 6'').
To solve the problem, our method has to maintain response naturalness while improving coherency of word choices.


\section{Conclusion}
We proposed a selection of response candidates generated from a neural conversational model (NCM) utilizing event causality relations.
The method had a robust matching of event causality relations attributed to distributed event representation.
Experimental results showed that the proposed method selects a coherent and diverse response.
The proposed method can be applied to any languages that have a semantic parser, because it uses predicate-argument structure based event expressions.
However, unnatural responses were sometimes selected due to inadequate event causality relations.
Future work will focus on solving the problem by preventing over-generalization of events, and maintaining response naturalness.


\section*{Acknowledgments}
We would like to thank Sadao Kurohashi, Ph.D. and Tomohide Shibata, Ph.D. of Kurohashi Laboratory in Kyoto University who provided us the event causality pairs.\par
This work is supported by JST PRESTO (JPMJPR165B).


\bibliography{acl2019}
\bibliographystyle{acl_natbib}



\end{document}